\documentclass[letterpaper, 10 pt, conference]{ieeeconf}  
\usepackage{graphicx}
\usepackage{amsmath}
\usepackage{amssymb}

\usepackage{amsthm}
\usepackage{url}
\usepackage{color}
\usepackage{multirow}
\usepackage{hhline}
\usepackage{amsfonts}
\usepackage{hyperref}
\usepackage{subfig}
\usepackage{bbm}
\let\labelindent\relax 
\usepackage{enumitem}

\newtheorem{remark}{Remark}

\newtheorem{problem}{Problem}

\IEEEoverridecommandlockouts                              
\usepackage{xcolor}

\title{\LARGE \bf
AuDeRe: Automated Strategy Decision and Realization in Robot Planning and Control via LLMs
}

\author{Yue Meng\textsuperscript{1,*}, Fei Chen\textsuperscript{1,*}, Yongchao Chen\textsuperscript{1,2}, and Chuchu Fan\textsuperscript{1}
\thanks{\textsuperscript{*}Equal contribution.}
\thanks{\textsuperscript{1}Massachusetts Institute of Technology. 
        mengyue@mit.edu, feic@mit.edu, chuchu@mit.edu.}%
\thanks{\textsuperscript{2}Harvard University. 
        yongchaochen@fas.harvard.edu.}        
}

\begin{document}

\maketitle
\thispagestyle{empty}
\pagestyle{empty}

\begin{abstract}
Recent advancements in large language models (LLMs) have shown significant promise in various domains, especially robotics. However, most prior LLM-based work in robotic applications either directly predicts waypoints or applies LLMs within fixed tool integration frameworks, offering limited flexibility in exploring and configuring solutions best suited to different tasks. In this work, we propose a framework that leverages LLMs to select appropriate planning and control strategies based on task descriptions, environmental constraints, and system dynamics. These strategies are then executed by calling the available comprehensive planning and control APIs. Our approach employs iterative LLM-based reasoning with performance feedback to refine the algorithm selection. We validate our approach through extensive experiments across tasks of varying complexity, from simple tracking to complex planning scenarios involving spatiotemporal constraints. The results demonstrate that using LLMs to determine planning and control strategies from natural language descriptions significantly enhances robotic autonomy while reducing the need for extensive manual tuning and expert knowledge. Furthermore, our framework maintains generalizability across different tasks and notably outperforms baseline methods that rely on LLMs for direct trajectory, control sequence, or code generation.
The source code can be found at: \url{https://github.com/mengyuest/llm-planning-control}.
\end{abstract}

\section{Introduction}
Recent progress in large language models has enabled the development of robotic planning systems that interpret and act on natural language instructions. These models have been applied to a variety of tasks in control and planning. Some approaches use LLMs to directly generate trajectory plans or even complete planner code, while others integrate them with existing tools to enhance decision-making. By basing planning on natural language inputs, these methods reduce reliance on specialized expertise and simplify the design process. This emerging paradigm paves the way for more intuitive and accessible robotic systems that can adapt their strategies based on high-level, human-readable descriptions.

Traditional planning and control algorithms provide strong theoretical guarantees when carefully chosen and tuned for a specific task. However, in complex and dynamic environments, selecting suitable strategies is a nontrivial challenge that requires significant expertise and fails to adapt to changes or scale effectively. To address this limitation, we propose a novel approach that leverages LLMs to select motion planning and control algorithms based on task descriptions in order to reduce human effort, improve adaptability, and enable broader applicability across diverse tasks.
Instead of directly generating trajectories or code, our framework uses LLMs to reason about task requirements, environmental constraints, and robot dynamics, subsequently deciding and invoking comprehensive planning and control application programming interfaces (APIs) tailored to these insights. We evaluate our method across diverse robotic scenarios with various complexity, ranging from simple trajectory tracking and basic planning tasks with collision avoidance to more complex scenarios involving maze navigation and high-level tasks encoded with signal temporal logic (STL)~\cite{maler2004monitoring} specifications. 
Our approach is benchmarked against two baselines: an end-to-end LLM prediction method (LLM-predict) that directly outputs trajectories or control sequences, and a method (LLM-code) where the LLM generates executable code. Performance is quantitatively assessed using success rates, the average number of query iterations required for successful task completion, iteration-based success rates, and a detailed analysis of different error types. Our results show that the proposed LLM-driven approach (LLM-use-API) significantly outperforms these baselines in terms of higher success rates, fewer iterative queries, and reduced occurrence of errors across all tested scenarios.
\begin{figure}[!h]
\centering
\includegraphics[width=\columnwidth]{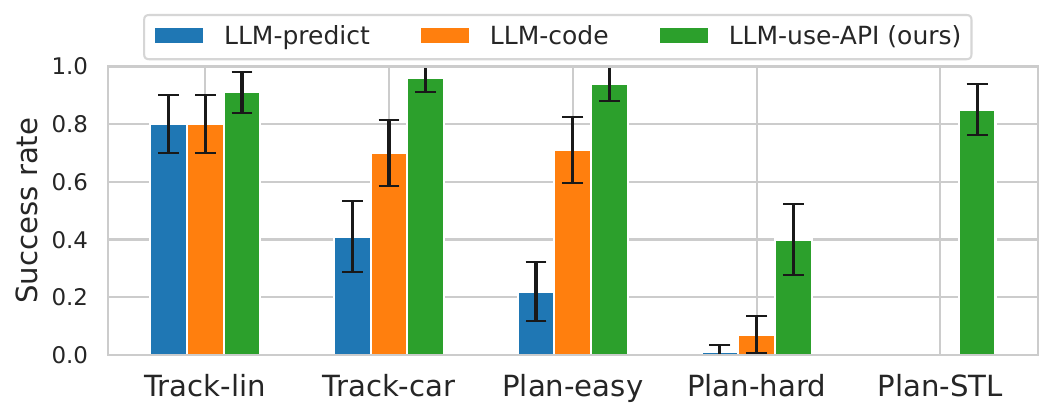}
\caption{LLM-use-API approach outperforms direct trajectory or code planners using LLMs.}
\label{teaser}
\end{figure}

\subsection{Literature review}   In~\cite{wang2024prompt}, the authors introduce a framework that uses few-shot prompts derived from physical environments. This framework allows LLMs to autoregressively predict low-level robotic control actions without requiring task-specific fine-tuning. An interface presented in~\cite{ismail2024narrate} integrates natural language instructions with underlying model predictive control module.  The work in~\cite{tagliabue2024real} exploits prior knowledge of UAV dynamics encoded within an LLM. It enables dynamic adaptation of the entire control stack by adjusting low-level parameters, optimizing trajectory tracking commands, and supporting decision-making at the mission planning level. The approach outlined in~\cite{wake2023chatgpt} employs customizable few-shot prompts to transform natural language instructions and environmental information into executable, multi-stage robotic task plans supported by iterative feedback mechanisms. In~\cite{liang2023code}, hierarchical prompting techniques are utilized to allow LLMs to autonomously generate robotic control code directly from natural language instructions. This structured approach effectively bridges semantic commands and executable robot actions. The integration in~\cite{ahn2022can} combines LLMs with pretrained robotic skill modules to ground semantic knowledge, which allows robots to interpret and execute complex natural language commands. The authors of~\cite{huang2022inner} demonstrate how closed-loop linguistic feedback significantly enhances the reasoning and planning capabilities of LLMs within embodied robotic applications. In~\cite{singh2023progprompt}, the authors propose a structured programmatic prompting method for robotic task planning that ensures reliable generation of action sequences adapted to specific environmental contexts. The methodology proposed in~\cite{ding2023task} combines LLMs with motion planning algorithms to achieve effective human-aligned multi-object rearrangement tasks utilizing common sense reasoning. Text2Motion, as described in~\cite{lin2023text2motion}, is a language-based robotic planning system that employs geometric feasibility heuristics and effectively solves complex sequential manipulation problems. The authors of~\cite{sharma2022correcting} propose the use of natural language instructions to refine robot goals and constraints, and show substantial improvements in planning effectiveness without reliance on real-time teleoperation. EUREKA, introduced in~\cite{ma2023eureka}, is an LLM-driven method for the design of reward functions tailored to complex manipulation tasks, and it clearly outperforms traditional human-crafted reward schemes across various reinforcement learning environments. The framework in~\cite{yu2023language} leverages LLMs to define effective reward structures and creates a connection between high-level language instructions and low-level robotic control actions. In~\cite{goyal2019using}, natural language instructions serve as a tool for reward shaping in reinforcement learning, and this approach leads to benefits in sample efficiency for complex Atari environments. Finally,~\cite{chen2024autotamp} introduces an approach based on few-shot translation and autoregressive re-prompting in LLMs to translate natural language commands into intermediate representations, which enables robust integration with task and motion planning modules for robotic operations.

\subsection{Statement of contributions}
In contrast to previous methods, our contributions are threefold. First, while lots of existing approaches either directly generate trajectories or rely solely on LLMs as code planners~\cite{chen2025code,liang2023code}, we utilize their advanced reasoning capabilities to intelligently select the appropriate planning and control strategies by considering detailed task descriptions, environmental constraints, and robotic system dynamics. Second, our method innovatively leverages LLMs for automated and context-aware algorithm selection rather than limiting integration to a single predefined tool. To the best of our knowledge, this is the first work to frame LLM-based robot planning and control as an automated strategy decision problem across multiple comprehensive planning and control APIs, rather than as direct trajectory generation or integration with a single fixed tool. This novel approach enhances flexibility and adaptability to diverse task requirements. Finally, we introduce an iterative feedback mechanism through performance-based re-prompting, enabling continuous refinement of strategy decisions. To validate our approach, we conduct extensive experiments across scenarios of varying complexity, demonstrating significant performance improvements and generalizability. All together, these contributions yield a powerful and versatile framework that advances the state-of-the-art in automated robot planning and control.


\section{Problem statement}
In this section, we introduce the problem setup, including system dynamics, objectives, and environmental constraints. Then, we formulate the problem of leveraging an LLM-based automated strategy for robot planning and control.

We consider robot planning and control tasks where high-level objectives and constraints are specified in natural language, while the robot's underlying dynamics are represented by the following continuous-time system:
\begin{equation}\label{eq:ds}
\dot{x}(t)=f(x,u),\quad x(t_0)=x_0\in \mathcal{X}_0,
\end{equation}
where $x(t)\in \mathcal{X} \subseteq \mathbb{R}^n$ denotes the states with $n$ indicating the spatial dimensions, $x_0\in \mathcal{X}_0$ represents the initial state and $\mathcal{X}_0\subset \mathcal{X}$, $u(t)\in \mathcal{U} \subseteq \mathbb{R}^m$ denotes the input vectors, the mapping $f: \mathbb{R}^n\times \mathbb{R}^m \rightarrow \mathbb{R}^n$ is assumed to be locally Lipschitz continuous with respect to $x$ and $u$.

We define the robot's workspace as $\mathcal{W} \subset \mathbb{R}^n$, and the tasks are categorized into planning, control, or a combination of both. For control tasks, the reference trajectory $s_r$ is either defined as a set of waypoints in discrete time or as a mapping $s_r: \mathbb{R}_{\geq 0} \rightarrow \mathbb{R}^n$ for a continuous-time trajectory. 
Regarding planning tasks, obstacles are represented by the set $\mathcal{O}:=\{{\mathcal{O}_1, \mathcal{O}_2, \ldots, \mathcal{O}_d}\}$, with each obstacle $\mathcal{O}_i \subset \mathcal{W}$ indicating regions that must be avoided. The collision-free workspace is therefore given by $\mathcal{W}_{\text{free}} = \mathcal{W} \setminus \bigcup_{i=1}^d \mathcal{O}_i$. The main objective is to navigate the robot from its initial state to the goal region $\mathcal{X}_g \subset \mathcal{W}_{\text{free}}$ while avoiding all obstacles. In addition to spatial constraints, the natural language task $\mathcal{T}$ may also capture certain temporal constraints, which can be further represented as a high-level STL task. Hence, $\mathcal{T}$ generally ranges from simple tracking control or planning to complex planning and even high-level task specifications under spatiotemporal constraints. Rather than explicitly planning trajectories or manually selecting an established planning and control strategy, we leverage the reasoning capabilities of large language models to automatically determine an appropriate strategy. This decision-making process relies on problem descriptions expressed in natural language, which are inherently more intuitive and accessible for humans.

The problem considered in this paper is how, given a clearly defined task $\mathcal{T}$ represented by natural language that encodes the robot dynamics, reference trajectory $s_r$, initial state $x_0$, goal region $\mathcal{X}_g$, and the environmental constraints $\mathcal{O}$, to effectively employ LLMs to select and refine a suitable subset of planning and control algorithms $\mathcal{A}^{\mathcal{T}}\subset \mathcal{A}$ from a comprehensive set of available methods set $\mathcal{A}$. Here, $\mathcal{A} = \{\mathcal{A}_1, \mathcal{A}_2, \dots, \mathcal{A}_N\}$ is a set of available planning and control APIs provided to the LLM. The planned trajectory or the closed-loop trajectory should satisfy the objectives encoded in $\mathcal{T}$ by automatically executing the selected algorithms $\mathcal{A}^{\mathcal{T}}$, particularly ensuring the robot reaches its goal region without collisions. Formally, we formulate our problem as follows: 
\begin{problem}
(LLM-based Strategy Decision.) 
Given a task $\mathcal{T}$ described in natural language, which encodes $(s_r, x_0, \mathcal{X}_g, \mathcal{O})$ and the dynamics~\eqref{eq:ds}, determine and refine a subset of algorithms $\mathcal{A}^{\mathcal{T}} \subseteq \mathcal{A}$ using an LLM such that the resulting planning or closed-loop trajectories satisfy $\mathcal{T}$. 
\end{problem}
\begin{remark}
The planning and control algorithms in $\mathcal{A}$ are composed of individual functions through APIs. It is nontrivial for an LLM to directly invoke these APIs due to the necessity of understanding detailed interactions, input/output relations, and ensuring compatibility in terms of data types and dimensions. Therefore, the LLM must first identify and configure the required parameters and interfaces, and explicitly generate integration code to seamlessly execute the selected APIs. 
\end{remark}

\section{LLM-based strategy decision}

In this section, we introduce an automated strategy decision process that leverages an LLM to select and iteratively refine planning and control strategies. First, we provide an overview of the approach, and then we detail each module.

\begin{figure}[!h]
\centering
\includegraphics[width=\columnwidth]{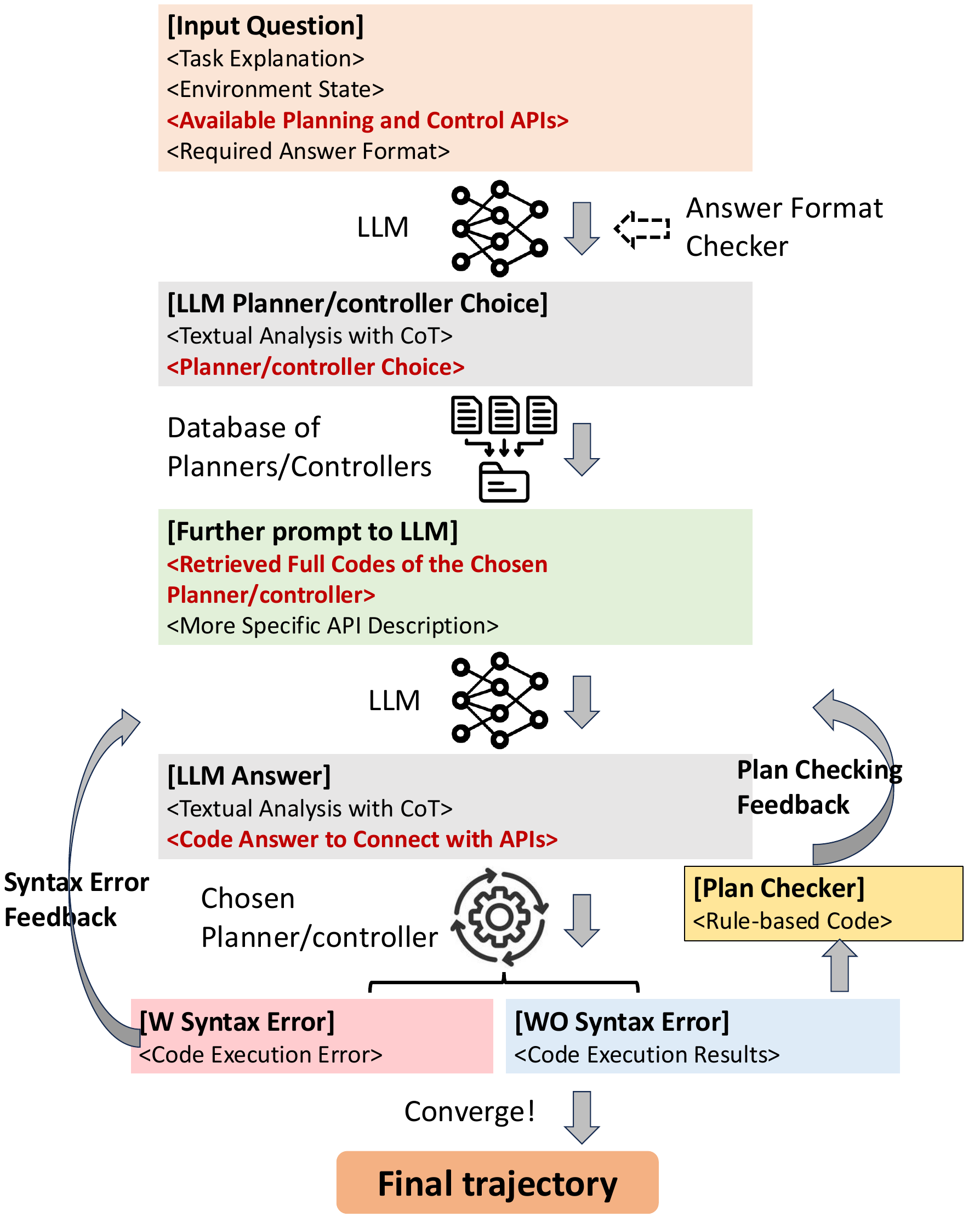}
\caption{The architecture for LLM-based strategy decision.}
\label{Control_LLM_Arch}
\end{figure}

\subsection{Approach in a nutshell}
The automated strategy decision process operates as follows and the illustration is shown in Fig. \ref{Control_LLM_Arch}: Given a task $\mathcal{T}$ described in natural language, which encodes an environment setup comprising reference trajectories, initial conditions, goal regions, environmental constraints, the robot's dynamics, and a set of available planning and control APIs denoted by $\mathcal{A} = \{\mathcal{A}_1, \mathcal{A}_2, \dots, \mathcal{A}_N\}$ is provided to the LLM. The LLM then selects an appropriate subset of these planning and control strategies, termed $\mathcal{A}^{\mathcal{T}} \subseteq \mathcal{A}$, to address the given task $\mathcal{T}$. Once the APIs are selected, the corresponding API codes are provided back to the LLM. At this stage, the LLM generates a high-level execution plan, explicitly describing how the selected APIs will interact, detailing the input/output relations, and ensuring that data types and dimensions match across interfaces. The LLM configures the necessary parameters and interfaces, and generates the integration code required to execute the selected APIs. If errors occur, such as syntax errors or planning failures, an iterative refinement loop is triggered with a predefined maximum number of iterations. Through these iterative improvements, the final strategy and execution either complete the task or, if the maximum iterations are reached, report the errors.

\subsection{Environment module} The environment module developed in this work provides a versatile and structured framework designed to facilitate the application and testing of LLMs for control and planning tasks. Specifically, it supports various dynamical systems, environmental constraints, initial states, and clearly defined target regions. The intended robot planning and control task, denoted by $\mathcal{T}$, is specified in natural language. The provided environmental functions and configurations comprehensively address the encoding and specification requirements associated with these problems formulated in natural language. 

The environment setup comprises multiple predefined dynamical models, including single and double integrators, unicycle, pendulum, and robotic-arm dynamics. These models allow comprehensive experiments across a range of planning and control scenarios. Each dynamical system is implemented across several computational frameworks such as PyTorch~\cite{paszke2019pytorch} for gradient-based control methods, and CasADi~\cite{andersson2019casadi} for optimization and model predictive control methods, providing flexibility for integration with different planning and control APIs. 

Environmental constraints are imposed through configurable obstacle types, including circles, squares, or mixed arrangements, while a boundary further defines the permissible workspace region $\mathcal{W}$. The module supports a broad range of task scenarios, spanning low-level trajectory tracking and basic state space discretization to region graph navigation, planning within these graphs, and ultimately tracking generated reference trajectories. Initial states and goal regions can be randomized within structured constraints to ensure diverse path planning challenges. These scenarios with different complexities enable comprehensive evaluations of the robustness and generalization capabilities of LLM-driven planning and control policy selection, and can be applied to much more complex high-level planning tasks involving temporal logic specifications. 

\subsection{Planning-control API module} 
The planning-control module developed in this work provides a comprehensive framework designed to integrate effectively with LLM-driven methods for a wide range of planning and control tasks. The module offers a flexible and intuitive interface to address scenarios ranging from low-level trajectory tracking, state space discretization to high-level planning involving complex temporal logic constraints.

To maximize versatility and effectiveness, the module provides eight distinct APIs, enabling an LLM to automatically select the appropriate approach based on specific problem settings and constraints. The provided APIs and their functionalities are summarized below. Our API set and task environments are chosen for clarity of evaluation, but the framework is not limited to these implementations. The APIs are modular and can be extended with additional planners, controllers, solvers, or external libraries. The LLM reasoning layer is environment-agnostic, so new algorithms can be added by exposing the relevant APIs. This design emphasizes the framework’s generalizability beyond the presented scenarios, as demonstrated in Sec.~\ref{subsec:scena}.
\begin{itemize}
    \item $A^\star$ search (astar)~\cite{hart1968formal}:  Performs graph-based optimal path planning using heuristic-driven search to efficiently navigate discrete state spaces and obstacles. 
    \item Cross entropy method (cem)~\cite{rubinstein2004cross}: Systematically explores a workspace, selects high-performing samples, and iteratively refines its search distribution to converge toward effective solutions. 
    \item Gradient-based optimization (grad): Employs gradient descent and backpropagation using PyTorch for trajectory optimization, ensuring efficient integration of differentiable dynamics and loss functions.
    \item Linear quadratic regulator (lqr)~\cite{anderson2007optimal}: Executes optimal tracking control for linearizable dynamical systems, enabling precise trajectory following by minimizing quadratic state and control costs.
    \item Mixed integer linear programming (milp): Provides robust trajectory planning under high-level STL specifications, which characterize both spatial and temporal constraints. 
    \item Model predictive control (mpc)~\cite{garcia1989model}: Provides optimal planning and control solutions over a prediction horizon using CasADi, supporting both linear and nonlinear dynamics with constraints.
    \item PID control (pid)~\cite{astrom1995pid}: Implements a straightforward yet effective proportional-integral-derivative control strategy for waypoint following tasks, which is suitable for simpler trajectory tracking scenarios.
    \item Rapidly-exploring random tree (rrt)~\cite{lavalle1998rapidly,karaman2011sampling}: Offers randomized path planning through either standard RRT or optimized RRT* algorithms, efficiently navigating high-dimensional or complex obstacle-rich spaces.
\end{itemize}

Each API is precisely defined with well-structured inputs and outputs, which allows efficient integration and straightforward functionality for LLMs. Their diverse nature enables LLMs to flexibly and effectively address a wide range of planning-control challenges. This provides strong generalization and adaptability across tasks of varying complexity.

\subsection{LLM module and feedback loop}  
The LLM module serves as a crucial interface between high-level natural language task descriptions and the detailed execution facilitated by the provided APIs. Initially, the input to the LLM consists of an environmental description, clearly defined task specifications, and a list of available planning-control APIs. Leveraging this context, the LLM decides the  appropriate APIs to fulfill the specified task. 

After the initial API selection, the LLM is prompted to retrieve detailed code for the corresponding APIs, including the required inputs, and expected outputs~\cite{lewis2020retrieval,retrieval2}. Using this comprehensive information, it then generates executable code snippets that interface with the selected APIs, configuring the necessary parameters and interfaces, and producing the integration code required for execution.
\begin{figure}[!h]
\centering
\includegraphics[width=\columnwidth, clip, trim=4.5cm 6.5cm 4.5cm 4.5cm]{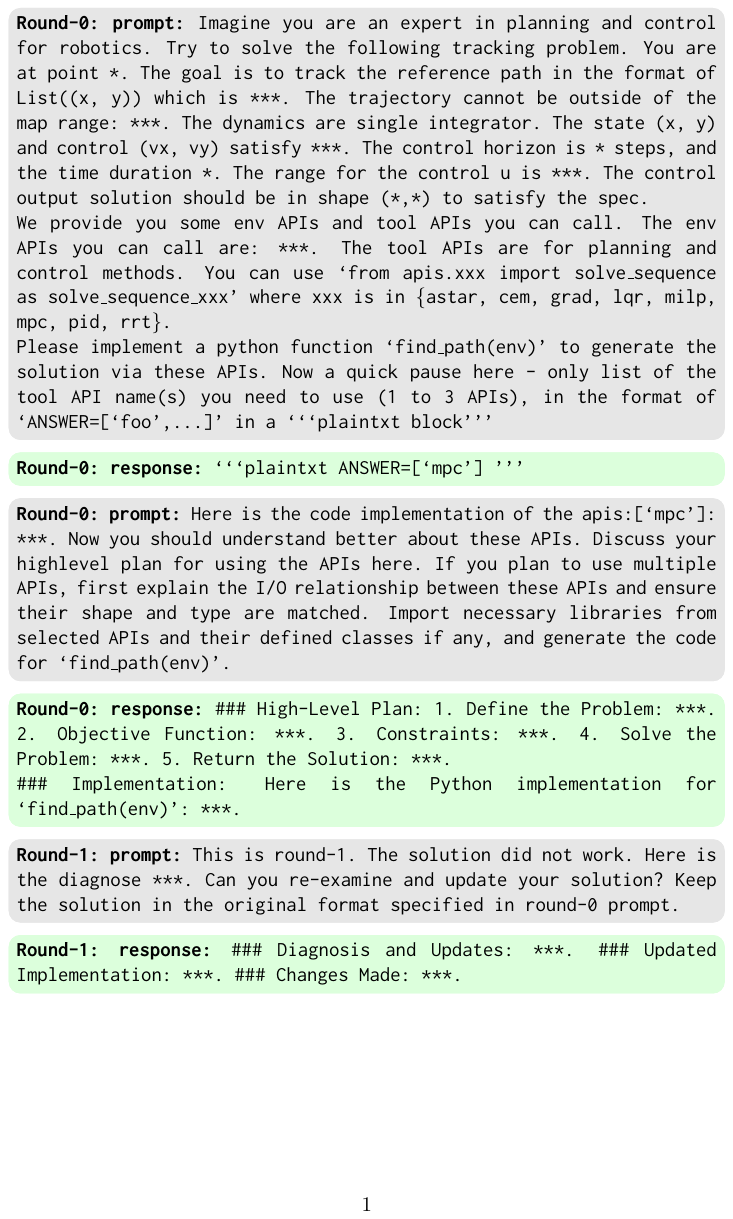}
\caption{Example prompt for a simple tracking problem with a single iteration.}
\label{prompt}
\end{figure}

The refinement process involves iterative feedback loops to ensure correctness and improved performance. Generated code undergoes syntax checking and planner checks to verify  compliance with specified goals, environmental constraints, and trajectories for tracking. Any spotted errors or implementations that fail to meet requirements trigger further iterative refinements by the LLM, in order to progressively improve the code quality and functional correctness~\cite{chen2024autotamp,corrective-reprompting}. We present an example on LLM prompt in Fig. \ref{prompt} for a simple tracking control problem, where the symbols $*$ or $***$ denote detailed parameters and text that can be omitted. In this example, we include the diagnostic summary from the previous round in the new prompt, and let LLM refine its strategy based on past performance feedback.

Through these iterative refinement cycles, which incorporate syntax validation and trajectory planning assessments, the final LLM-based strategy converges to reliable planning and closed-loop trajectories that accurately and effectively fulfill the tasks. The entire LLM-based strategy decision process and the roles of LLMs are illustrated by Fig. \ref{Control_LLM_Arch}.

\section{Experiments}
In this section, we present the experimental setup designed to evaluate the performance of our proposed method. We first describe the scenarios considered, outline the baseline methods selected for comparison, and detail the metrics used to assess the performance. Finally, we demonstrate through extensive experiments results that our method consistently outperforms these baselines across various scenarios.


\begin{figure}[!h]
  \centering
  \subfloat[Simple tracking with linear dynamics.]{%
    \includegraphics[width=0.45\columnwidth]{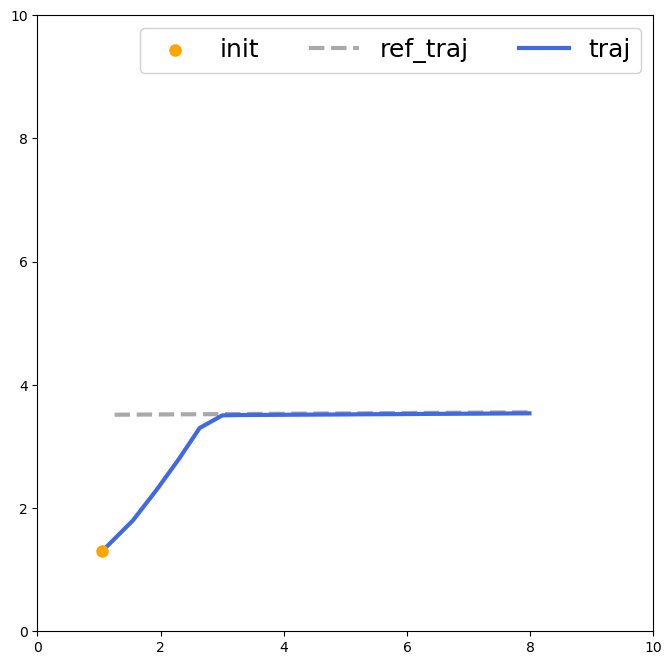}%
    \label{fig:sub1}%
  }
  \hfill
  \subfloat[Simple tracking with nonlinear Dubins car dynamics.]{%
    \includegraphics[width=0.45\columnwidth]{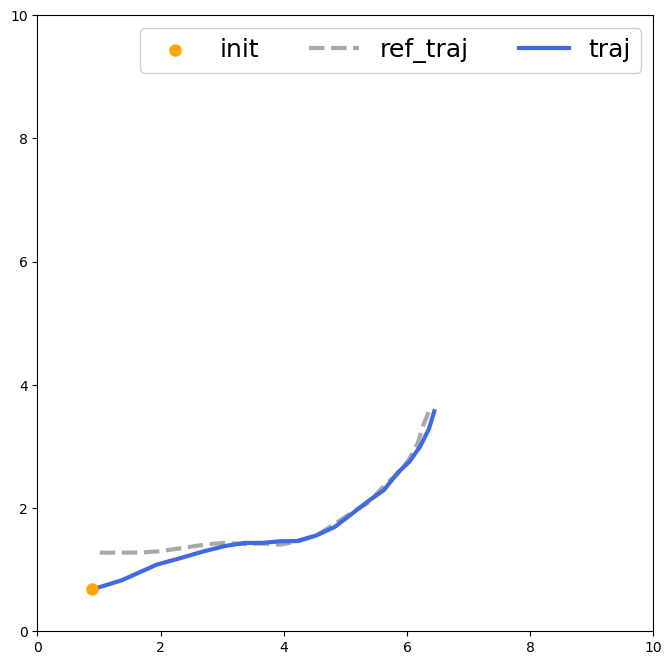}%
    \label{fig:sub2}%
  }
  
  \vskip\baselineskip
  
  \subfloat[Simple planning with collision avoidance.]{%
    \includegraphics[width=0.45\columnwidth]{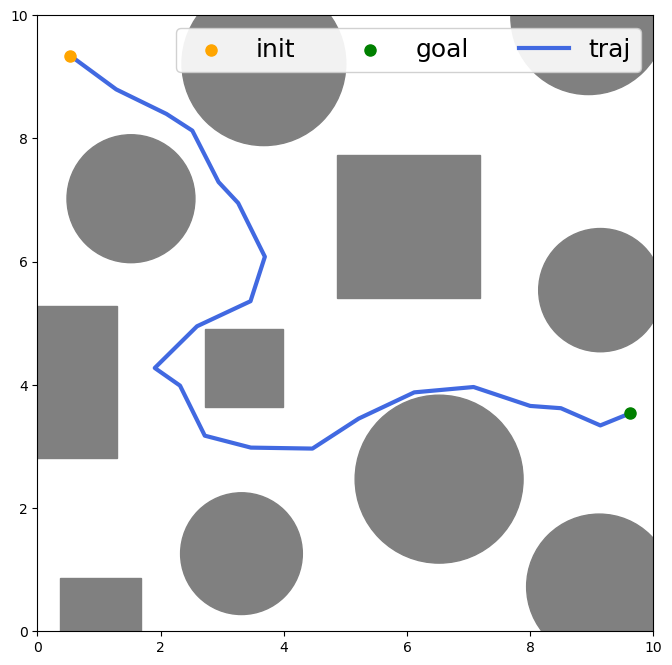}%
    \label{fig:sub3}%
  }
  \hfill
  \subfloat[Complex planning in a $3 \times 3$ maze with extensive collisions.]{%
    \includegraphics[width=0.45\columnwidth]{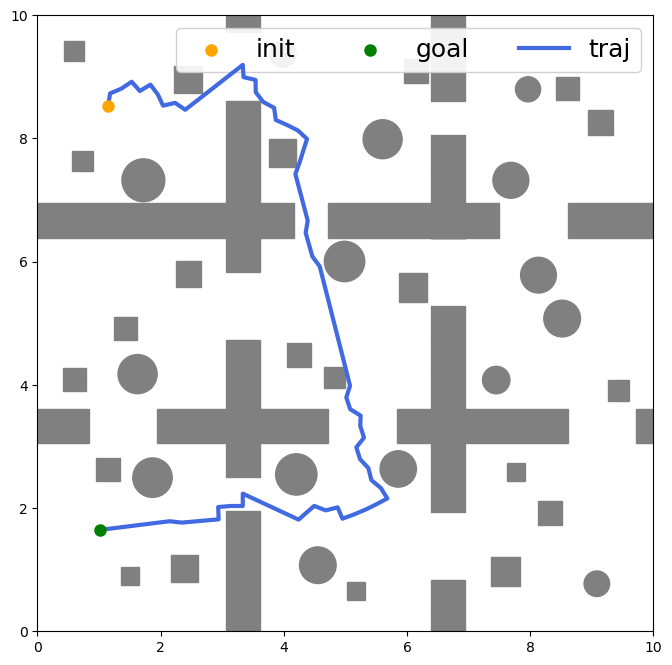}%
    \label{fig:sub4}%
  }
  
  \vskip\baselineskip
  
  \subfloat[High-level task under STL specifications.]{%
    \includegraphics[width=0.45\columnwidth]{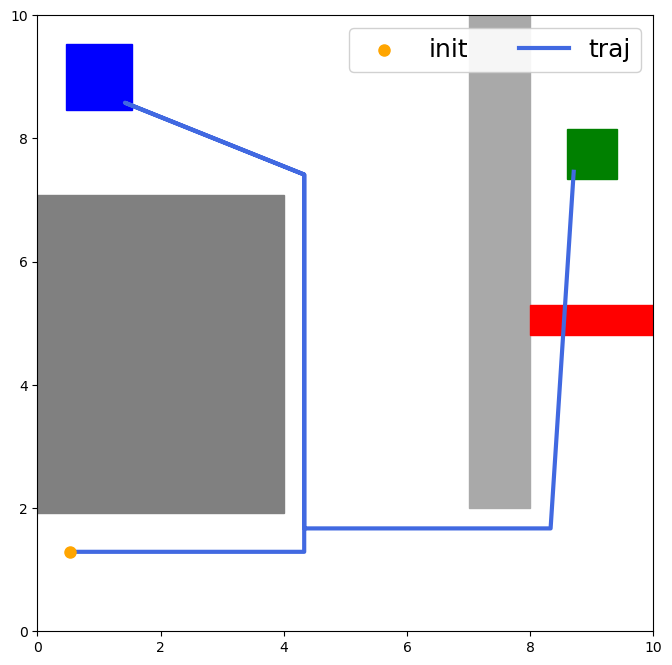}%
    \label{fig:sub5}%
  }
  
  \caption{Robot planning and control scenarios with different complexities.}
  \label{fig:combined}
\end{figure}

\begin{figure*}[h]
    \centering
    \includegraphics[width=\textwidth]{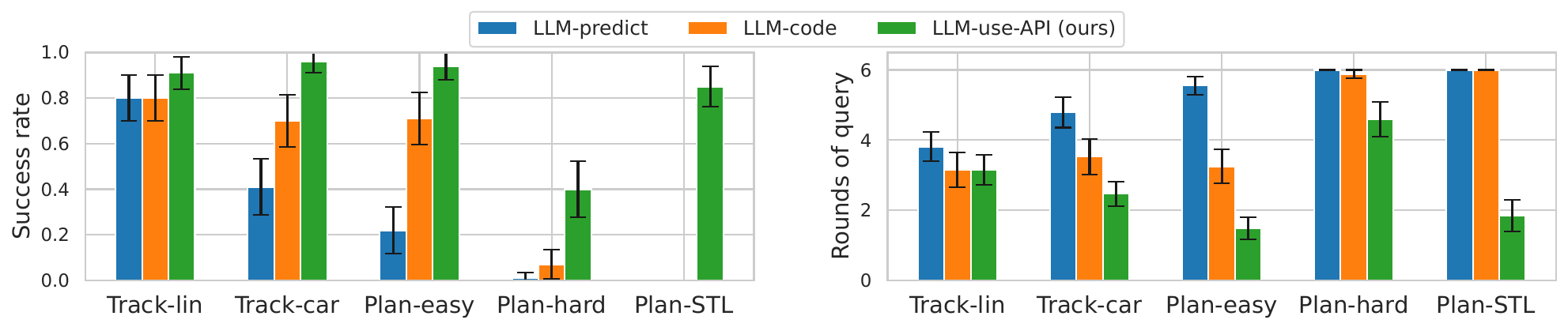} 
    \caption{Comparison of success rates (left) and number of query rounds (right) across various LLM-based approaches with varying task complexities.}
    \label{fig4}
\end{figure*}
\begin{figure*}[h]
    \centering
    \includegraphics[width=0.9\textwidth]{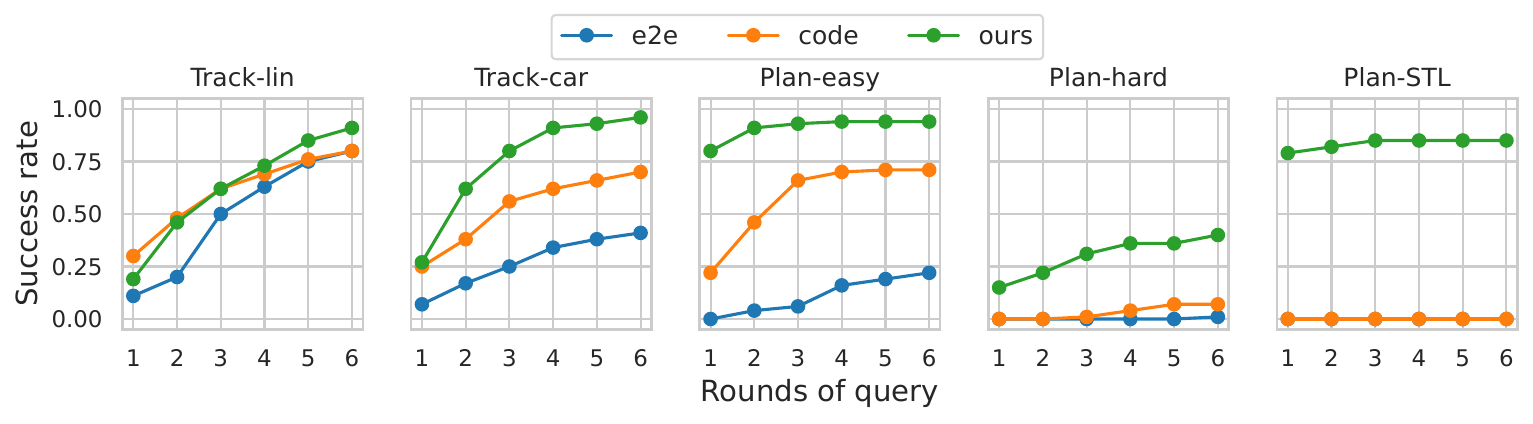} 
    \caption{Comparison of the success rate of various LLM-based approaches as the number of query rounds increases.}
    \label{fig5}
\end{figure*}
\begin{figure*}[h]
    \centering
    \includegraphics[width=\textwidth]{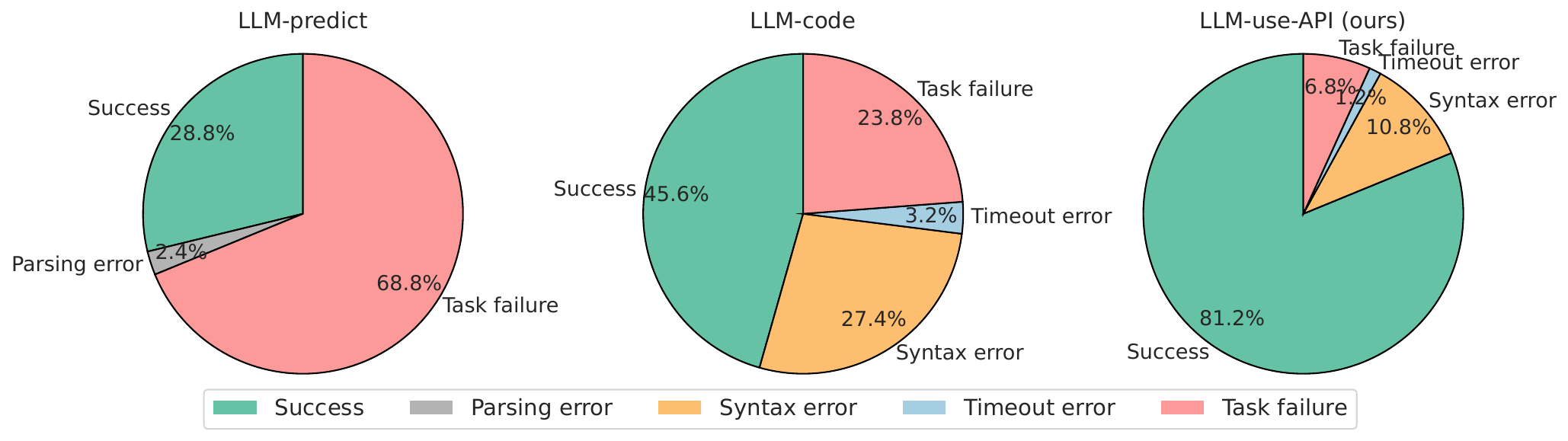} 
    \caption{Statistics of the error types for different LLM-based approaches.}
    \label{fig6}
\end{figure*}

\subsection{Scenarios}\label{subsec:scena}
We consider five representative robot planning and control scenarios that increase in complexity. These range from basic trajectory tracking and simple planning to complex planning and high-level tasks involving sophisticated spatial and temporal constraints. Specifically, the evaluated scenarios are as follows and as shown in Fig. \ref{fig:combined}:
\begin{enumerate}[label=\alph*)]
    \item Simple tracking with linear dynamics: A basic tracking task involving linear system dynamics, aiming to accurately follow a given reference trajectory.
    \item Simple tracking with nonlinear Dubins car dynamics: A trajectory tracking scenario involving a nonlinear Dubins car model.
    \item Simple planning with collision avoidance: A planning task that requires navigation from an initial state to a goal region while avoiding collisions with randomly placed obstacles.
    \item Complex planning in a $3\times 3$ maze with extensive collision environment: A more intricate planning scenario where the robot must efficiently navigate through a structured $3\times 3$ maze while avoiding multiple obstacles distributed throughout the environment.
    \item High-level task under STL specifications: A sophisticated scenario incorporating STL specifications that encode spatial and temporal constraints. The robot is tasked with picking up a key, unlocking and entering a room, and subsequently reaching a goal region while satisfying specific spatiotemporal constraints. 
\end{enumerate}
For generalizability, each scenario is extensively evaluated over $100$ experiments. We randomize environmental parameters such as reference trajectories, initial states, goal regions, and obstacle placements, ensuring diverse and representative conditions for a thorough comparison. Fig. \ref{fig:combined} shows one successful experiment per scenario. In the simple tracking problems (a) and (b), the ``mpc" API is called. For the simple planning scenario (c), the ``rrt" API is employed. In the complex $3\times 3$ maze planning case, a combination of the ``astar" and ``rrt" APIs is applied. For the high-level STL task in (e), the ``milp" API is invoked. All experiments are implemented using the GPT-4o large language model.

\subsection{Baselines} 
We consider two baseline approaches for comparison with our proposed method. The first baseline is an \textbf{end-to-end (e2e)} approach, referred to as \textbf{LLM-predict}, where the LLM directly predicts the entire control sequence or planning trajectories from the problem description without external computational assistance. The second baseline, denoted as \textbf{LLM-code}, employs the LLM directly generating executable Python code tailored to the specified tasks~\cite{chen2025code,chen2025codesteer}. As discussed in the literature review, these baselines capture the two dominant paradigms of prior LLM-based planning: direct trajectory generation and code generation. Other approaches are typically designed for specific tasks or rely on additional learning frameworks, and thus address different problem formulations outside the scope of our evaluation.

Our method, labeled \textbf{LLM-use-API}, leverages the strengths of LLM to determine effective planning and control strategies while assigning the computational execution of these strategies to specialized APIs. We systematically compare LLM-use-API against both baseline methods across all previously defined scenarios to evaluate the corresponding performance and effectiveness.

\subsection{Performance metrics}
To evaluate and compare each method (LLM-predict, LLM-code, and our LLM-use-API), we conduct 100 experiments per scenario. Each method is allowed a maximum of six iterative rounds of queries or reprompting in case errors occurred. We collect three key metrics: 1) the success rate, which is the percentage of experiments in which the method successfully completed the task; 2) the average number of iterations to success, representing the average number of query rounds or repromptings needed to achieve a successful outcome; and 3) the iteration-based success rate that shows the success rate as a function of the number of iterations, which provides insights into the efficiency of each method.

Additionally, we monitor and categorize errors encountered during the experiments into four main types for comprehensive analysis: 1) Parsing errors, such as unsuccessful function loading or incorrect parsing of instructions; 2) Syntax errors, which arise from runtime issues due to syntactical mistakes in the generated code; 3) Timeout errors, occurring when code execution exceeds predefined maximum time limits; and 4) Task failures, where the generated code runs without errors but fails to produce valid control sequences or planning trajectories that fulfill the given tasks. These errors are logged systematically to facilitate a detailed performance comparison among the evaluated methods.

\subsection{Experiment results}
The experimental results are shown in Figs. \ref{fig4}, \ref{fig5} and \ref{fig6}  based on the performance metrics described previously. 

Fig. \ref{fig4} presents a comparison of success rates (left) and the average number of query rounds (right) for various LLM-based approaches across tasks of different complexity. We observe that all three methods (LLM-predict, LLM-code, and LLM-use-API) achieve high success rate on simpler tasks. However, for more complex scenarios such as extensive collision planning tasks or tasks involving high-level spatiotemporal constraints, the performance of LLM-predict and LLM-code drops significantly, and often fail completely. In contrast, our LLM-use-API method consistently outperforms these baselines across all difficulty levels, demonstrating higher success rate and requiring fewer rounds of queries to achieve successful task completion. 

Fig. \ref{fig5} illustrates the comparison of success rate as the number of query iterations increases. It clearly shows that for all approaches, increased query iterations or feedback rounds lead to higher success rate. Notably, our LLM-use-API method consistently exhibits higher efficiency by reaching higher success rate with fewer iterations compared to the baselines. 

Fig. \ref{fig6} summarizes the statistics of different error types encountered across the evaluated LLM-based approaches. Our LLM-use-API approach significantly reduces errors across all categories, including parsing errors, syntax errors, timeout errors, and task failures, resulting in a considerably higher overall success rate. 

These results demonstrate that while simpler tasks can be effectively addressed through direct trajectory prediction or code generation using LLMs, our proposed method, which systematically leverages external computational APIs, outperforms traditional LLM approaches that rely solely on direct trajectories or code planners in complex robot planning and control tasks.

\subsection{Ablation study}
\begin{figure}[!h]
\centering
\includegraphics[width=\columnwidth]{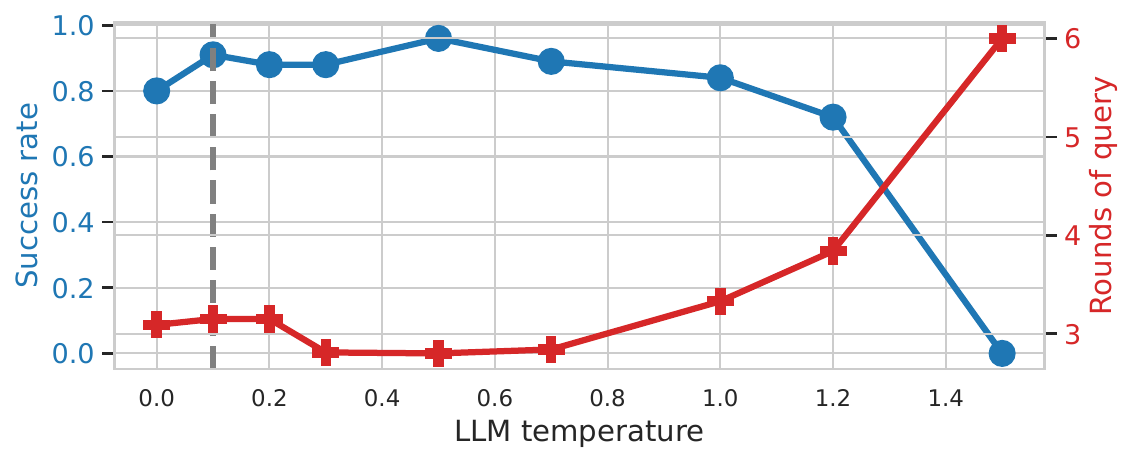}
\caption{Performance comparison of various LLM temperature settings using GPT-4o.}
\label{abl1}
\end{figure}

We further conduct an ablation study to evaluate the impact of different large language models and temperature parameter settings on our LLM-use-API framework’s performance. We first focus on the effect of the temperature parameter in our LLM-use-API framework, specifically using the GPT-4o model. The temperature parameter in an LLM controls the randomness of generated outputs: lower temperatures produce more deterministic results, while higher temperatures introduce greater variations and randomness~\cite{temperature-effect}. We analyze the impact of temperature settings within a complex planning scenario that usually combines $A^\star$ and RRT algorithms. The results, as shown in Fig. \ref{abl1}, reveal that optimal performance occurs with temperature values between 0.1 and 0.7, which is consistent with our earlier experiments conducted at a temperature of 0.1. At higher temperature settings, we observe significant performance degradation. 
\begin{figure}[!h]
\centering
\includegraphics[width=\columnwidth]{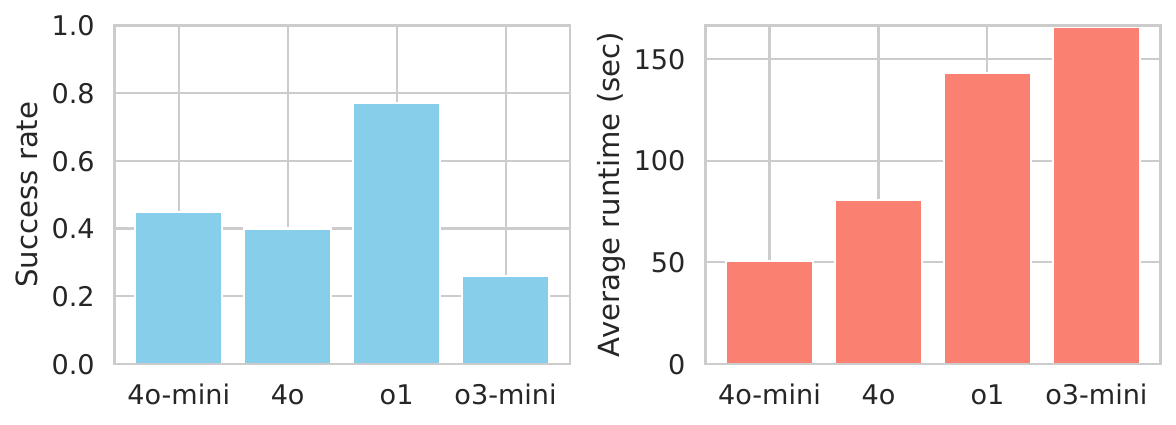}
\caption{Comparative performance analysis of various large language models.}
\label{abl2}
\end{figure}
\begin{figure}[!h]
\centering
\includegraphics[width=\columnwidth]{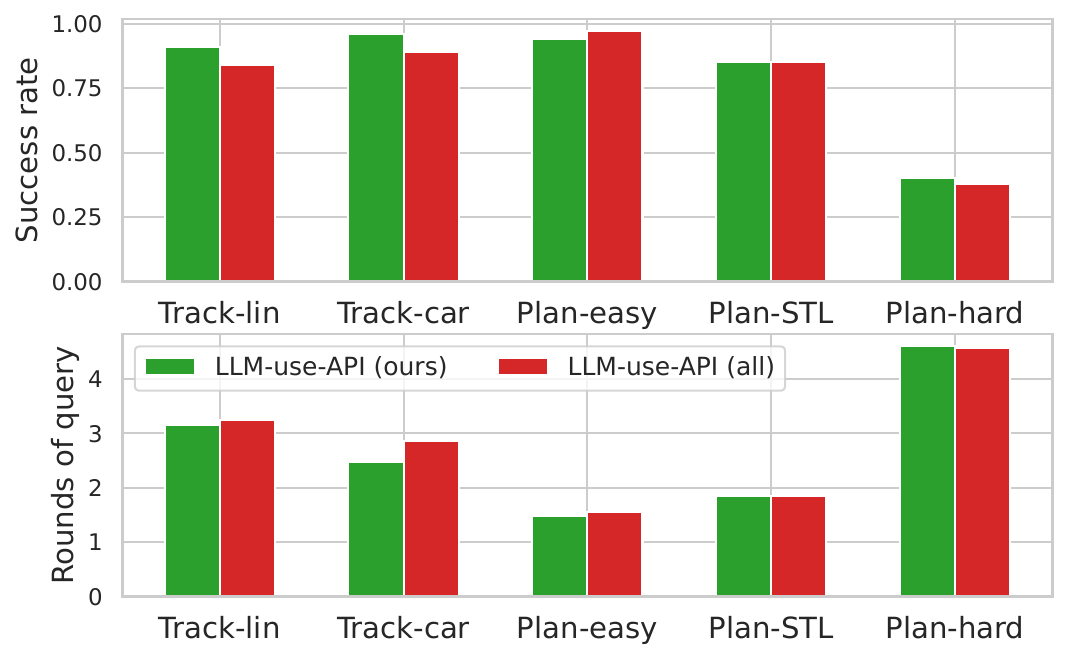}
\caption{Ablation study on the impact of API code provision strategy.}
\label{abl3}
\end{figure}
On the other hand, we compare the performance of several recent LLM variants, including GPT-4o-mini, o1, and o3-mini, in the same complex planning scenario. The results as shown in Fig. \ref{abl2}, indicate that o1 outperforms the other evaluated models for the complex planning tasks. Even the newer o3-mini exhibits surprisingly lower performance. Additionally, larger models incur higher computational runtimes.

As a further ablation study, we compare our method, which provides API code to the LLM only after the corresponding APIs are called, with an alternative approach where all available API codes are sent directly to the LLM at once. The results shown in Fig. \ref{abl3} indicate that the two methods perform similarly, with our approach achieving a slightly higher success rate and fewer iterative query rounds. By exposing only the relevant APIs, our method is more efficient and cost-effective in both API usage and prompt design.

\section{Conclusions} 
In this work, we explored leveraging large language models for automated algorithm selection in robotic motion planning and control. Unlike traditional methods that directly predict trajectories or generate code, or those integrating a single fixed tool, our proposed framework intelligently selects appropriate planning and control strategies based on task descriptions, environmental constraints, and system dynamics. Experimental evaluations across tasks of varying complexity demonstrate that our method significantly outperforms direct LLM prediction and code generation baselines, and achieves higher success rates with fewer errors and query iterations. We have also conducted an ablation study to evaluate the impact of different large language models and temperature parameter settings on our LLM-use-API framework’s performance. Future directions include extending this approach to more complex tasks, evaluating more large language models, and validating performance through hardware experiments or detailed simulator demonstrations.


\bibliographystyle{abbrv}        
\bibliography{mybib}

\begin{thebibliography}{10}

\bibitem{ahn2022can}
M.~Ahn, A.~Brohan, N.~Brown, Y.~Chebotar, O.~Cortes, B.~David, C.~Finn, C.~Fu, K.~Gopalakrishnan, K.~Hausman, et~al.
\newblock Do as i can, not as i say: Grounding language in robotic affordances.
\newblock {\em arXiv preprint arXiv:2204.01691}, 2022.

\bibitem{anderson2007optimal}
B.~D. Anderson and J.~B. Moore.
\newblock {\em Optimal control: linear quadratic methods}.
\newblock Courier Corporation, 2007.

\bibitem{andersson2019casadi}
J.~A. Andersson, J.~Gillis, G.~Horn, J.~B. Rawlings, and M.~Diehl.
\newblock Casadi: a software framework for nonlinear optimization and optimal control.
\newblock {\em Mathematical Programming Computation}, 11:1--36, 2019.

\bibitem{chen2024autotamp}
Y.~Chen, J.~Arkin, C.~Dawson, Y.~Zhang, N.~Roy, and C.~Fan.
\newblock Autotamp: Autoregressive task and motion planning with llms as translators and checkers.
\newblock In {\em 2024 IEEE International conference on robotics and automation (ICRA)}, pages 6695--6702. IEEE, 2024.

\bibitem{chen2025codesteer}
Y.~Chen, Y.~Hao, Y.~Liu, Y.~Zhang, and C.~Fan.
\newblock Codesteer: Symbolic-augmented language models via code/text guidance.
\newblock {\em arXiv preprint arXiv:2502.04350}, 2025.

\bibitem{chen2025code}
Y.~Chen, Y.~Hao, Y.~Zhang, and C.~Fan.
\newblock Code-as-symbolic-planner: Foundation model-based robot planning via symbolic code generation.
\newblock {\em arXiv preprint arXiv:2503.01700}, 2025.

\bibitem{ding2023task}
Y.~Ding, X.~Zhang, C.~Paxton, and S.~Zhang.
\newblock Task and motion planning with large language models for object rearrangement.
\newblock In {\em 2023 IEEE/RSJ International Conference on Intelligent Robots and Systems (IROS)}, pages 2086--2092. IEEE, 2023.

\bibitem{garcia1989model}
C.~E. Garcia, D.~M. Prett, and M.~Morari.
\newblock Model predictive control: Theory and practice—a survey.
\newblock {\em Automatica}, 25(3):335--348, 1989.

\bibitem{goyal2019using}
P.~Goyal, S.~Niekum, and R.~J. Mooney.
\newblock Using natural language for reward shaping in reinforcement learning.
\newblock {\em arXiv preprint arXiv:1903.02020}, 2019.

\bibitem{hart1968formal}
P.~E. Hart, N.~J. Nilsson, and B.~Raphael.
\newblock A formal basis for the heuristic determination of minimum cost paths.
\newblock {\em IEEE transactions on Systems Science and Cybernetics}, 4(2):100--107, 1968.

\bibitem{huang2022inner}
W.~Huang, F.~Xia, T.~Xiao, H.~Chan, J.~Liang, P.~Florence, A.~Zeng, J.~Tompson, I.~Mordatch, Y.~Chebotar, et~al.
\newblock Inner monologue: Embodied reasoning through planning with language models.
\newblock {\em arXiv preprint arXiv:2207.05608}, 2022.

\bibitem{ismail2024narrate}
S.~Ismail, A.~Arbues, R.~Cotterell, R.~Zurbr{\"u}gg, and C.~A. Alonso.
\newblock Narrate: Versatile language architecture for optimal control in robotics.
\newblock In {\em 2024 IEEE/RSJ International Conference on Intelligent Robots and Systems (IROS)}, pages 9628--9635. IEEE, 2024.

\bibitem{karaman2011sampling}
S.~Karaman and E.~Frazzoli.
\newblock Sampling-based algorithms for optimal motion planning.
\newblock {\em The international journal of robotics research}, 30(7):846--894, 2011.

\bibitem{lavalle1998rapidly}
S.~LaValle.
\newblock Rapidly-exploring random trees: A new tool for path planning.
\newblock {\em Research Report 9811}, 1998.

\bibitem{lewis2020retrieval}
P.~Lewis, E.~Perez, A.~Piktus, F.~Petroni, V.~Karpukhin, N.~Goyal, H.~K{\"u}ttler, M.~Lewis, W.-t. Yih, T.~Rockt{\"a}schel, et~al.
\newblock Retrieval-augmented generation for knowledge-intensive nlp tasks.
\newblock {\em Advances in neural information processing systems}, 33:9459--9474, 2020.

\bibitem{liang2023code}
J.~Liang, W.~Huang, F.~Xia, P.~Xu, K.~Hausman, B.~Ichter, P.~Florence, and A.~Zeng.
\newblock Code as policies: Language model programs for embodied control.
\newblock In {\em 2023 IEEE International Conference on Robotics and Automation (ICRA)}, pages 9493--9500. IEEE, 2023.

\bibitem{lin2023text2motion}
K.~Lin, C.~Agia, T.~Migimatsu, M.~Pavone, and J.~Bohg.
\newblock Text2motion: From natural language instructions to feasible plans.
\newblock {\em Autonomous Robots}, 47(8):1345--1365, 2023.

\bibitem{ma2023eureka}
Y.~J. Ma, W.~Liang, G.~Wang, D.-A. Huang, O.~Bastani, D.~Jayaraman, Y.~Zhu, L.~Fan, and A.~Anandkumar.
\newblock Eureka: Human-level reward design via coding large language models.
\newblock {\em arXiv preprint arXiv:2310.12931}, 2023.

\bibitem{maler2004monitoring}
O.~Maler and D.~Nickovic.
\newblock Monitoring temporal properties of continuous signals.
\newblock In {\em International Symposium on Formal Techniques in Real-Time and Fault-Tolerant Systems}, pages 152--166. Springer, 2004.

\bibitem{paszke2019pytorch}
A.~Paszke.
\newblock Pytorch: An imperative style, high-performance deep learning library.
\newblock {\em arXiv preprint arXiv:1912.01703}, 2019.

\bibitem{retrieval2}
O.~Ram, Y.~Levine, I.~Dalmedigos, D.~Muhlgay, A.~Shashua, K.~Leyton-Brown, and Y.~Shoham.
\newblock In-context retrieval-augmented language models.
\newblock {\em Transactions of the Association for Computational Linguistics}, 11:1316--1331, 2023.

\bibitem{corrective-reprompting}
S.~S. Raman, V.~Cohen, E.~Rosen, I.~Idrees, D.~Paulius, and S.~Tellex.
\newblock Planning with large language models via corrective re-prompting.
\newblock In {\em NeurIPS 2022 Foundation Models for Decision Making Workshop}, 2022.

\bibitem{temperature-effect}
M.~Renze.
\newblock The effect of sampling temperature on problem solving in large language models.
\newblock In {\em Findings of the Association for Computational Linguistics: EMNLP 2024}, pages 7346--7356, 2024.

\bibitem{rubinstein2004cross}
R.~Y. Rubinstein and D.~P. Kroese.
\newblock {\em The cross-entropy method: a unified approach to combinatorial optimization, Monte-Carlo simulation and machine learning}.
\newblock Springer Science \& Business Media, 2004.

\bibitem{sharma2022correcting}
P.~Sharma, B.~Sundaralingam, V.~Blukis, C.~Paxton, T.~Hermans, A.~Torralba, J.~Andreas, and D.~Fox.
\newblock Correcting robot plans with natural language feedback.
\newblock {\em arXiv preprint arXiv:2204.05186}, 2022.

\bibitem{singh2023progprompt}
I.~Singh, V.~Blukis, A.~Mousavian, A.~Goyal, D.~Xu, J.~Tremblay, D.~Fox, J.~Thomason, and A.~Garg.
\newblock Progprompt: Generating situated robot task plans using large language models.
\newblock In {\em 2023 IEEE International Conference on Robotics and Automation (ICRA)}, pages 11523--11530. IEEE, 2023.

\bibitem{tagliabue2024real}
A.~Tagliabue, K.~Kondo, T.~Zhao, M.~Peterson, C.~T. Tewari, and J.~P. How.
\newblock Real: Resilience and adaptation using large language models on autonomous aerial robots.
\newblock In {\em 2024 IEEE 63rd Conference on Decision and Control (CDC)}, pages 1539--1546. IEEE, 2024.

\bibitem{wake2023chatgpt}
N.~Wake, A.~Kanehira, K.~Sasabuchi, J.~Takamatsu, and K.~Ikeuchi.
\newblock Chatgpt empowered long-step robot control in various environments: A case application.
\newblock {\em IEEE Access}, 11:95060--95078, 2023.

\bibitem{wang2024prompt}
Y.-J. Wang, B.~Zhang, J.~Chen, and K.~Sreenath.
\newblock Prompt a robot to walk with large language models.
\newblock In {\em 2024 IEEE 63rd Conference on Decision and Control (CDC)}, pages 1531--1538. IEEE, 2024.

\bibitem{yu2023language}
W.~Yu, N.~Gileadi, C.~Fu, S.~Kirmani, K.-H. Lee, M.~G. Arenas, H.-T.~L. Chiang, T.~Erez, L.~Hasenclever, J.~Humplik, et~al.
\newblock Language to rewards for robotic skill synthesis.
\newblock {\em arXiv preprint arXiv:2306.08647}, 2023.

\bibitem{astrom1995pid}
K.~J. Åström and T.~Hägglund.
\newblock {\em PID Controllers: Theory, Design, and Tuning}.
\newblock Instrument Society of America, Research Triangle Park, NC, 1995.

\end{thebibliography}

\end{document}